\documentclass{article}

\usepackage{microtype}
\usepackage{graphicx}
\usepackage{subfigure}
\usepackage{booktabs} 
\usepackage[section]{placeins}

\usepackage{hyperref}

\usepackage[accepted]{icml2020}

\usepackage{amsmath,amsfonts,bm}

\def\eqref#1{equation~\ref{#1}}

\def\1{\bm{1}}

\def\vv{{\bm{v}}}

\def\vx{{\bm{x}}}

\def\vz{{\bm{z}}}

\def\mA{{\bm{A}}}

\DeclareMathAlphabet{\mathsfit}{\encodingdefault}{\sfdefault}{m}{sl}
\SetMathAlphabet{\mathsfit}{bold}{\encodingdefault}{\sfdefault}{bx}{n}

\def\gN{{\mathcal{N}}}

\newcommand{\R}{\mathbb{R}}

\DeclareMathOperator*{\argmax}{arg\,max}

\usepackage{url}
\usepackage{cleveref}
\usepackage{amsmath}
\usepackage{amssymb}
\usepackage{changepage}

\icmltitlerunning{On Predictive Information in RNNs}

\begin{document}

\twocolumn[
\icmltitle{On Predictive Information in RNNs}




\icmlsetsymbol{equal}{*}
\begin{icmlauthorlist}
\icmlauthor{Zhe Dong}{goog}
\icmlauthor{Deniz Oktay}{goog,princeton}
\icmlauthor{Ben Poole}{goog}
\icmlauthor{Alexander A. Alemi}{goog}
\end{icmlauthorlist}

\icmlcorrespondingauthor{Zhe Dong}{zhedong@google.com}
\icmlcorrespondingauthor{Alexander A. Alemi}{alemi@google.com}
\icmlaffiliation{goog}{Google Research, Mountain View, California, USA}
\icmlaffiliation{princeton}{Princeton University, Princeton, New Jersey, USA}

\icmlkeywords{Machine Learning, Information Theory}

\vskip 0.3in
]
\printAffiliationsAndNotice{}

\begin{abstract}
Certain biological neurons demonstrate a remarkable capability to optimally compress the history of sensory inputs while being maximally informative about the future. In this work, we investigate if the same can be said of artificial neurons in recurrent neural networks (RNNs) trained with maximum likelihood. Empirically, we find that RNNs are suboptimal in the information plane. Instead of optimally compressing past information, they extract additional information that is not relevant for predicting the future. We show that constraining past information by injecting noise into the hidden state can improve RNNs in several ways: optimality in the predictive information plane, sample quality, heldout likelihood, and downstream classification performance.
\end{abstract}

\section{Introduction}

Remembering past events is a critical component of predicting the future and acting in the world. An information-theoretic quantification of how much observing the past can help in predicting the future is given by the \emph{predictive information}~\citep{predictive}. The predictive information is the mutual information (MI) between a finite set of observations (the \emph{past} of a sequence) and an infinite number of additional draws from the same process (the \emph{future} of a sequence). As a mutual information, the predictive information gives us a reparameterization independent, symmetric, interpretable measure of the co-dependence of two random variables.  More colloquially, the mutual information tells us how many \emph{bits} we can predict of the future given our observations of the past. Asymptotically, a vanishing fraction of the information in the past is relevant to the future~\citep{predictive}, thus systems which excel at prediction need not memorize the entire past of a sequence.

Intriguingly, certain biological neurons extract representations that efficiently capture the predictive information in sequential stimuli~\citep{palmer2015predictive, tkavcik2016information}. In~\citet{palmer2015predictive}, spiking responses of neurons in salamander retina had near optimal mutual information with the \emph{future} states of sequential stimuli they were exposed to, while compressing the \emph{past} as much as possible.  

Do artificial neural networks perform similarly? In this work, we aim to answer this question by measuring artificial recurrent neural networks' (RNNs) ability to compress the past while retaining relevant information about the future. 

Our contributions are as follows:
\begin{itemize}
    \item We demonstrate that RNNs, unlike biological systems, are suboptimal at extracting predictive information on the tractable sequential stimuli used in \citet{palmer2015predictive}. 
    \item We thoroughly validate the accuracy of our mutual information estimates on RNNs and optimal models, highlighting the importance of heldout sets for mutual information estimation.
    \item We show that RNNs trained with constrained capacity representations are closer to optimal on simple sequential stimuli and sketch datasets, and can improve sample quality, diversity, heldout log-likelihood, and downstream classification performance on several real-world sketch datasets \citep{ha2017sketchrnn} in the limited data regime.
\end{itemize}

\section{Background and Methods}
We begin by providing additional background on predictive information,  mutual information estimators, stochastic RNNs, and the Gaussian Information Bottleneck. These tools are necessary for accurately evaluating the question of whether RNNs are optimal in the information plane, as we require knowledge of the optimal frontier, and accurate estimates of mutual information for complex RNN models.

\subsection{Predictive Information}

Imagine an infinite sequence of data $(\dots, X_{t-1}, X_t, X_{t+1}, \dots$).  The predictive information \citep{predictive} of the sequence is the mutual information between some finite number of observations of the past ($T$) and the infinite future of the sequence:
\begin{align*}
I_{\text{pred}}(T) & = I(X_{\text{past}}; X_{\text{future}}) \\
& = I(\{X_{t-T+1}, \dots, X_{t}\}; \{ X_{t+1}, \dots \}).
\end{align*}
For a process for which the dynamics are not varying in time, this will be independent of the particular time $t$ chosen to be the present. More specifically, the predictive information is an expected log-ratio between the likelihood of observing a future given the past and observing that future in expectation over all possible pasts:
\begin{align*}
    I_{\text{pred}} & \equiv
    \mathbb{E}_{p(x_{\text{past}}, x_{\text{future}})} \left[ \log \frac{p(x_{\text{future}} | x_{\text{past}}) }{p(x_{\text{future}})}  \right] \\
    & = \mathbb{E}_{p(x_{\text{past}}, x_{\text{future}})}\left[ \log \frac{p(x_{\text{future}} | x_{\text{past}}) }{\mathbb{E}_{p(x_{\text{past}}')} \left[ p(x_{\text{future}} | x_{\text{past}}') \right]}  \right]. 
\end{align*} 
 
A sequential model such as an RNN provides a stochastic representation of the entire past of the sequence $Z \sim p(z|x_{\text{past}})$. For any such representation, we can measure how much information it retains about the past, a.k.a. the \emph{past information}: $I_{\text{past}} = I(Z; X_{\text{past}})$, and how informative it contains about the future, a.k.a. the \emph{future information}: $I_{\text{future}} = I(Z;X_{\text{future}})$. Because the representation depends only on the past, our three random variables satisfy the Markov relations: $Z \leftarrow X_{\text{past}} \leftrightarrow X_{\text{future}}$ and the Data Processing Inequality \citep{coverthomas} ensures that the information we have about the future is always less than or equal to both the true predictive information of the sequence ($I_\text{future} \leq I_{\text{pred}}$) as well as the information we retain about the past ($I_{\text{future}} \leq I_{\text{past}}$). For any particular sequence, there will be a frontier of solutions that optimally tradeoff between $I_\text{past}$ and $I_\text{future}$. A common method for tracing out this frontier is through the Information Bottleneck Lagrangian~\citep{DBLP:journals/corr/physics-0004057}:
\begin{equation}
    \min_{p(z|x_\text{past})}\; I(Z; X_\text{past}) - \beta I(Z; X_\text{future}),
    \label{eqn:iblag}
\end{equation}
where the parameter $\beta$ controls the tradeoff. An \emph{efficient} representation of the past is one that lies on this \emph{optimal frontier}, or equivalently is a solution to Eqn.~\ref{eqn:iblag} for a particular choice of $\beta$. For simple problems, where the sequence is jointly Gaussian, we will see that the optimal frontier can be identified analytically.

\subsection{Mutual Information estimators}

In order to measure whether a representation is efficient, we need a way to measure its past and future informations. While mutual information estimation is difficult in general~\citep{paninski2003estimation,mcallester2019formal}, recent progress has been made on a wide range of variational bounds on mutual information~\citep{alemi2016deep,poole2019variational}. While these provide bounds and not exact estimates of mutual information, they allow us to compare mutual information quantities in continuous spaces across models. There are two broad families of estimators: variational lower bounds powered by a tractable generative model, or contrastive lower bounds powered by an unnormalized critic.

The former class of lower bounds, first presented in \citet{barber2003algorithm}, are powered by a variational generative model:
\begin{align*}
    I_\text{future} & = \mathbb{E}_{p(x_\text{past}, x_\text{future})p(z|x_\text{past})}\left[ \log \frac{p(x_{\text{future}} | z)}{p(x_{\text{future}})} \right] \\ & \geq H(x_{\text{future}}) + \mathbb{E}_{p(x_\text{past}, x_\text{future})p(z|x_\text{past})}\left[ \log q(x_{\text{future}}| z) \right].
\end{align*}
A generative model provides a demonstration that there exists at least some information between the representation $z$ and the future of the sequence. For our purposes, $H(x_\text{future})$ (the entropy of the future of the sequence), is a constant determined by the dynamics of the sequence itself and outside our control.  For tractable problems, such as the toy problem we investigate below, this value is known. For real datasets, this value is not known, so we cannot produce reliable estimates of the mutual information. It does, however, still provide reliable gradients of a lower bound on $I_\text{future}$. One example of such a generative model is the loss used to train the RNN to begin with.  

Contrastive lower bounds can be used to estimate $I_\text{future}$ for datasets where building a tractable generative model of the future is challenging. InfoNCE style lower bounds~\citep{oord2018representation,poole2019variational} only require access to \emph{samples} from both the joint distribution and the individual marginals:
\begin{align}
&I(X; Z) \ge I_\text{NCE}(X; Z) \nonumber\\
&\triangleq  \mathbb{E}_{p^K(x,z)}\left[\frac{1}{K}\sum_{i=1}^K \log \frac{e^{f(x_i, z_i)}}{\frac{1}{K}\sum_{j=1}^K e^{f(x_j,z_i)}}\right].
\label{eqn:infonce}
\end{align}
Here $f(x_j, z_i)$ is a trained \emph{critic} that plays a role similar to the discriminator in a Generative Adversarial Network~\citep{ganpaper}.  It scores pairs, attempting to determine if an $(x,z)$ pair came from the joint ($p(x,z)$) or the factorized marginal distributions ($p(x)p(z)$).

When forming estimates of $I_\text{past}$, we can leverage additional knowledge about the known encoding distribution from the stochastic RNN $p(z|x_{\text{past}})$ to form tractable upper and lower bounds without having to learn an additional critic \citep{poole2019variational}: 
\begin{align}
&\mathbb{E}\left[ \frac 1 K \sum_{i=1}^K \log \frac{p(z^i|x_{\text{past}}^i)}{\frac 1 K \sum_{j} p(z^i | x_{\text{past}}^j )}  \right] 
\leq I(Z;X_{\text{past}}) \nonumber \\
&\leq 
\mathbb{E}\left[ \frac 1 K \sum_{i=1}^K \log \frac{p(z^i|x_{\text{past}}^i)}{\frac 1 {K-1} \sum_{j\neq i} p(z^i | x_{\text{past}}^j )}  \right] .
\label{eq:mb}
\end{align}
We refer to these bounds as \emph{minibatch upper and lower bounds} as they are computed using minibatches of size $K$ from the dataset. As the minibatch size $K$ increases, the upper and lower bounds can become tight. When $\log K \ll I(Z;X_{\mathrm{past}})$ the lower bound saturates at $\log K$ and the upper bound can be loose, thus we require using large batch sizes to form accurate estimates of $I_\text{past}$.

\subsection{Constraining information with stochastic RNNs}
\label{sec:stochastic_rnn}
\begin{figure}[t!]
    \centering
    \includegraphics[width=0.7\linewidth]{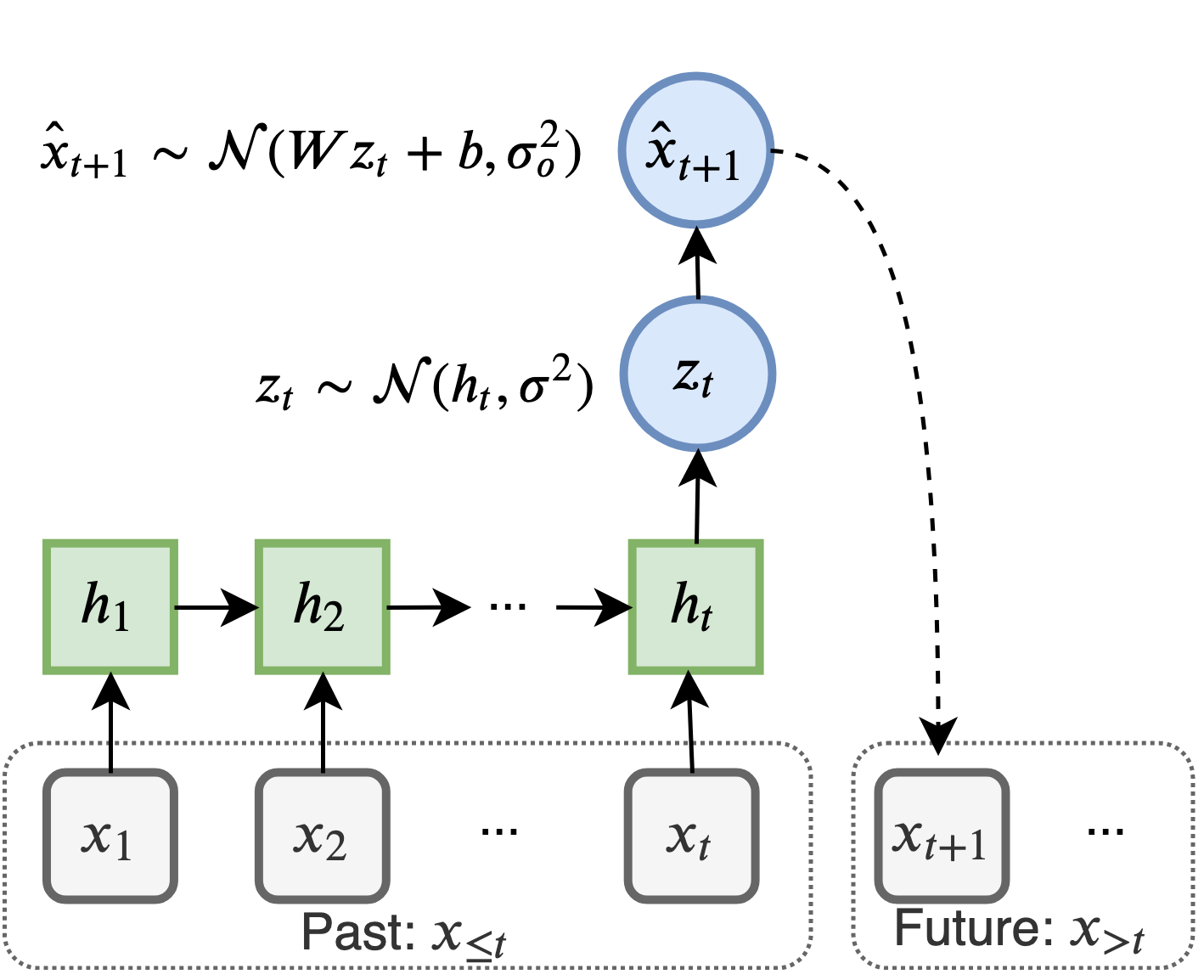}
    \caption{
    Schematic of Gaussian-noise-augmented stochastic RNN. 
    } 
    \label{fig:stochastic_rnn}
\end{figure}

Deterministic RNNs can theoretically encode infinite information about the past in their hidden states (up to floating point precision). To limit past information, we devise a simple stochastic RNN. Given the deterministic hidden state $h_t$, we output a stochastic variable $z_t$ by adding i.i.d. Gaussian noise to the hidden state before reading out the outputs: $z_t \sim \mathcal{N}(h_t, \sigma^2)$. We refer to $\sigma$ as the ``noise level'' for the stochastic RNN. These stochastic outputs are then used to predict the future state: $\hat{x}_{t+1} \sim \mathcal{N}(\text{g}_\text{decoder}(z_t), \sigma_o^2)$, as illustrated in~\Cref{fig:stochastic_rnn}. With bounded activation functions on the hidden state $h_t$, we can use $\sigma^2$ to upper bound the information stored about the past in the stochastic latent $z_t$. This choice of stochastic recurrent model yields a tractable conditional distribution $p(z_t|x_{\le t}) \sim \mathcal{N}(h_t, \sigma^2)$, which we can use in Eq.~\ref{eq:mb} to form tractable upper and lower bounds on the past information. We will consider two different settings for our stochastic RNNs: (1) where the RNNs are trained deterministically and the noise on the hidden state is added only at evaluation time,  and (2) where the RNNs are trained with noise, and evaluated with noise. We refer to the former as deterministically trained RNNs, and the latter as constrained information RNNs or RNNs trained with noise.

\subsection{Gaussian Information Bottleneck}
\label{sec:gib}

To evaluate optimality of RNNs, we first focus on the tractable sequential stimuli of a simple Brownian Harmonic Oscillator as used in \citep{palmer2015predictive}. A crucial property of this dataset is that we can analytically calculate the optimal trade-off between past and future information as it is an instance of the Gaussian Information Bottleneck~\citep{chechik2005gib}.

Consider jointly multivariate Gaussian random variables $X \in \R^{D_X}$ and $Y \in \R^{D_Y}$, with covariance $\Sigma_X$ and $\Sigma_Y$ and cross-covariance $\Sigma_{XY}$. The solution to the Information Bottleneck objective:
\begin{equation}
    \min_{T} I(X;T) - \beta I(Y;T),
\end{equation}
is given by a linear transformation $T = \mA X + \varepsilon$ with $\varepsilon \sim \gN(\mathbf{0}, \Sigma_\varepsilon)$. The projection matrix $A$ projects along the lowest eigenvectors of $\Sigma_{X|Y} \Sigma_X^{-1}$, where the trade-off parameter $\beta$ decides how many of the eigenvectors participate. Further details can be found in \Cref{app:gib}.

\section{Experimental Results}

\begin{figure}[t!]
\centering
\includegraphics[width=0.6\linewidth]{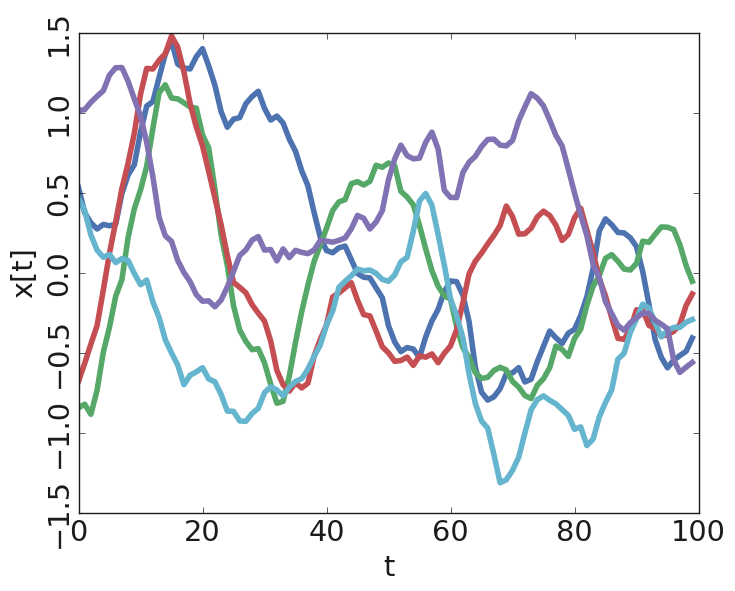}
\caption{Example trajectories of Brownian harmonic oscillator over time, with each color representing a different trajectory.}
\label{fig:bho_inputs}
\end{figure}

\begin{figure*}[h!]
\centering
\includegraphics[width=\linewidth]{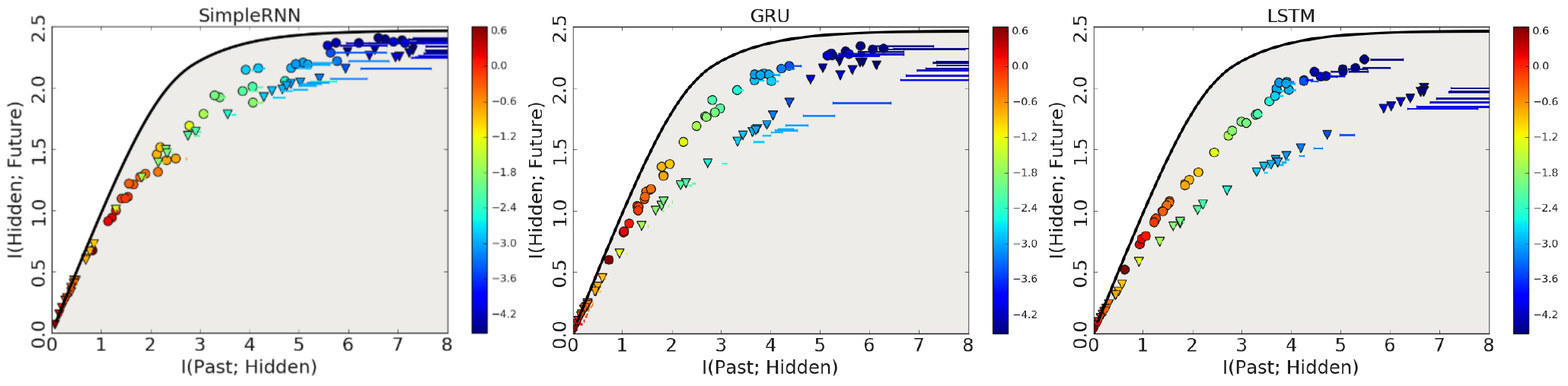}
\caption{
Estimates of the past information contained in the stochastic variable (x-axis) vs. future information (y-axis). The feasible region is shaded. The models to the upper-left perform better future prediction, while compressing the past more, therefore they are of higher efficiency. Points correspond to estimates with a learned critic; bars represent the gap between the lower and upper bounds using the known conditional distribution instead of a learned critic; colors represent the noise level, $\log$(noise), added to the hidden state. The stochastically trained RNN is marked as $\circ$, and deterministically trained RNN with post-hoc noise injection is marked as $\triangledown$.}
\label{fig:bho}
\end{figure*}

To compare the efficiency of RNNs at extracting predictive information to biological systems, we begin with experiments on data sampled from a Brownian harmonic oscillator (BHO), matching the stimuli displayed to neurons in salamander retina in \citet{palmer2015predictive}. Samples from a BHO are
a form of Ornstein–Uhlenbeck process.
This system has the benefit of having analytically tractable predictive information.
The dynamics are given by:
\begin{align}
    \begin{split}
        x_{t+\Delta t} ~ = & ~ x_t + v_t \Delta t , \\
        v_{t+\Delta t} ~ = & ~[ 1 - \Gamma \Delta t] v_t - \omega^2 x_t \Delta t + \xi_t \sqrt{D \Delta t}.
    \end{split}
    \label{eqn:bho_eom}
\end{align}
where $\xi_t$ is a standard Gaussian random variable. Additional details can be found in \Cref{app:bho_training_details}.
Examples of the trajectories for this system can be seen in~\Cref{fig:bho_inputs}.

\subsection{Are RNNs efficient in the information plane?}
Given its analytical tractability we can explicitly assess the estimated 
RNN performance against optimal performance. 
We compared three major variants of RNNs, including fully-connected RNNs, gated recurrent units (GRU,~\citet{cho2014gru}), and LSTMs~\citep{lstm}.
Each network had $32$ hidden units and {\tt tanh} activations. Full training details
are in \Cref{app:bho_training_details}. 

By training the RNNs without noise and injecting noise to RNN hidden states at evaluation time,
one can produce networks with compressed representations.
By varying the strength of the
noise, networks trace out a trajectory
on the information plane.
We find that these deterministically trained networks 
with noise added post-hoc leave considerable gaps between the information frontier of the model (colored $\triangledown$s) and the optimal frontier (black),
as demonstrated in \Cref{fig:bho}.

\subsection{Are information constrained RNNs more efficient on the information plane?}
\label{sec:bho_training}
We find that networks trained with noise injection are more efficient at capturing predictive information than networks trained deterministically but evaluated with post-hoc noise injection.
In \Cref{fig:bho}, comparing the results for 
stochastic RNNs 
and deterministic RNNs with noise added only at
evaluation time, 
We find that networks trained with noise are close to the optimal frontier (black), nearly optimal at extracting information about the past that is useful for predicting the future.
While injecting noise at evaluation time produces
networks with compressed representations, these
deterministically trained networks perform worse than their
stochastically trained counterparts.

At the same noise level (the color coding in \Cref{fig:bho})
stochastically trained RNNs have
both higher $I(\vz; \vx_{\text{past}})$ and higher $I(\vz; \vx_{\text{future}})$.
By limiting the capacity of the models during training, constrained information RNNs are able to extract more efficient representations.
For this task, we surprisingly find that
more complex RNN variants such as LSTMs are less efficient
at encoding predictive information. This may be due to optimization difficulties \citep{collins2016capacity}.

\subsection{How sensitive are these findings to MI estimator and training objectives?}
Our claim that RNNs are suboptimal in capturing predictive information hinges 
on the quality of our MI estimates, and may also be impacted by the choice of training objectives beyond 
maximum likelihood estimation. Are the RNNs truly suboptimal or is it just that our MI estimates or modeling choices are inappropriate? Here we provide several experiments further validating our claims.

\begin{figure}[!htb]
    \centering
    \includegraphics[width=0.6\linewidth]{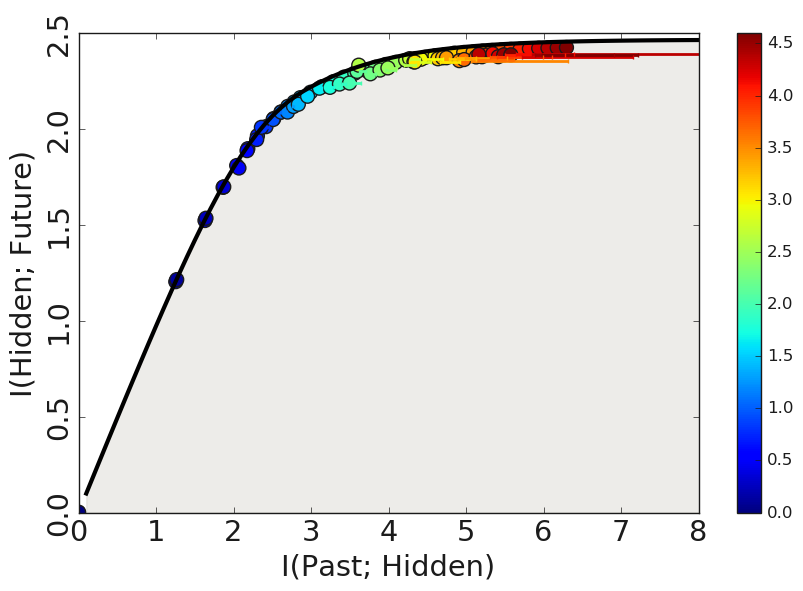}
    \caption{{\bf Evaluating mutual information estimators given optimal encoder}. For past information estimation, I(Hidden; Past), InfoNCE lower bound (colored points) and MB-Lower and MB-Upper (colored bars) are used; for future information, I(Hidden; Future), only InfoNCE is applied, due to the lack of a tractable conditional distribution for $p(y_t|z_t)$. Color represents the level of the trade-off parameter $\beta$ in the IB Lagrangian.}
    \label{fig:gib_theory}
\end{figure}
\begin{figure}[!htb]
\centering
\includegraphics[width=0.65\linewidth]{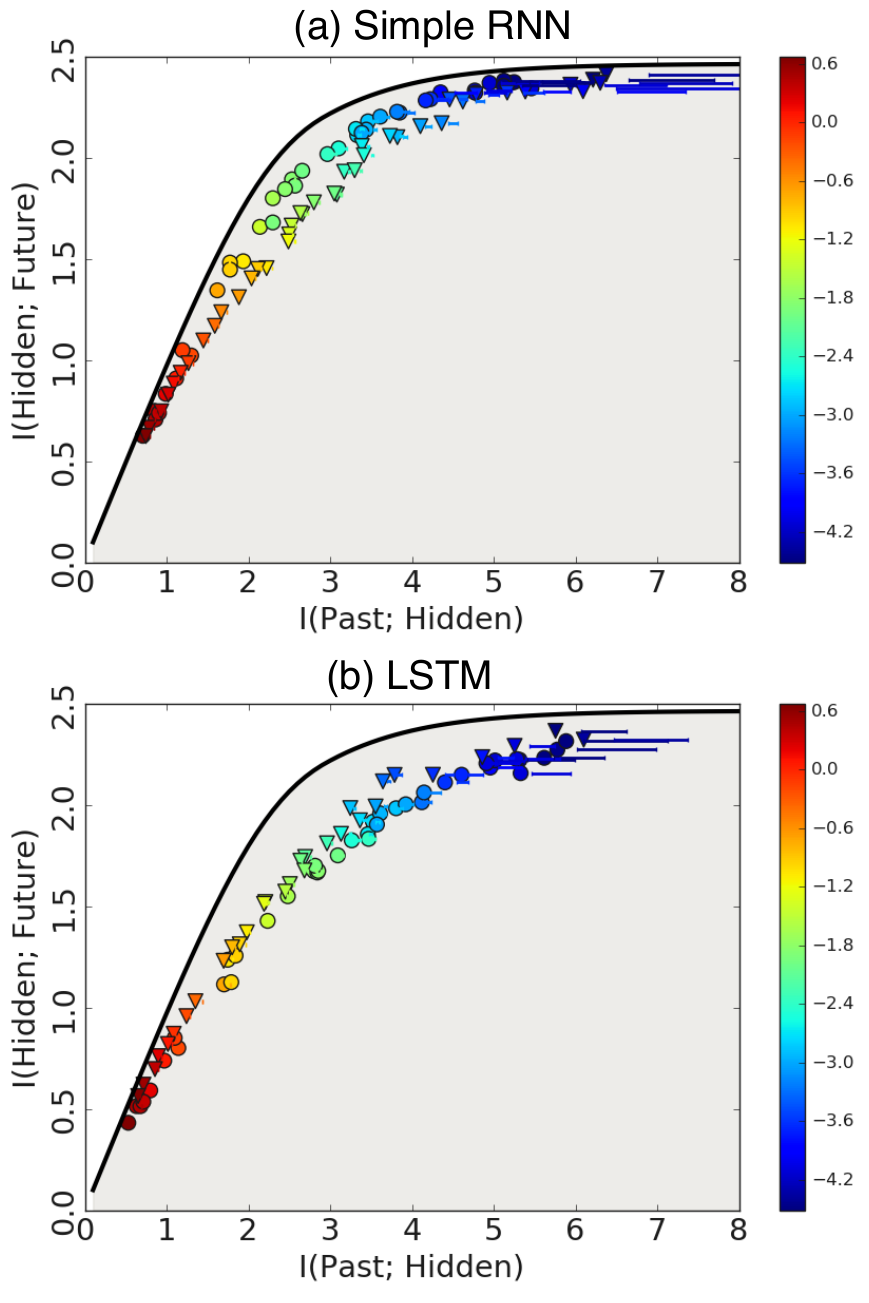}
\caption{The impact of training objectives on BHO dataset for fully connected RNN {\bf (top)} and LSTM {\bf (bottom)}. Models trained with maximum likelihood estimations are marked with $\circ$, and models trained with the contrastive loss are marked with $\triangledown$. The color bar shows the noise level in $\log_{10}$ scale.}
\label{fig:app_cpc_vs_mle}
\end{figure}
\begin{figure}[!htb]
\centering
\includegraphics[width=0.65\linewidth]{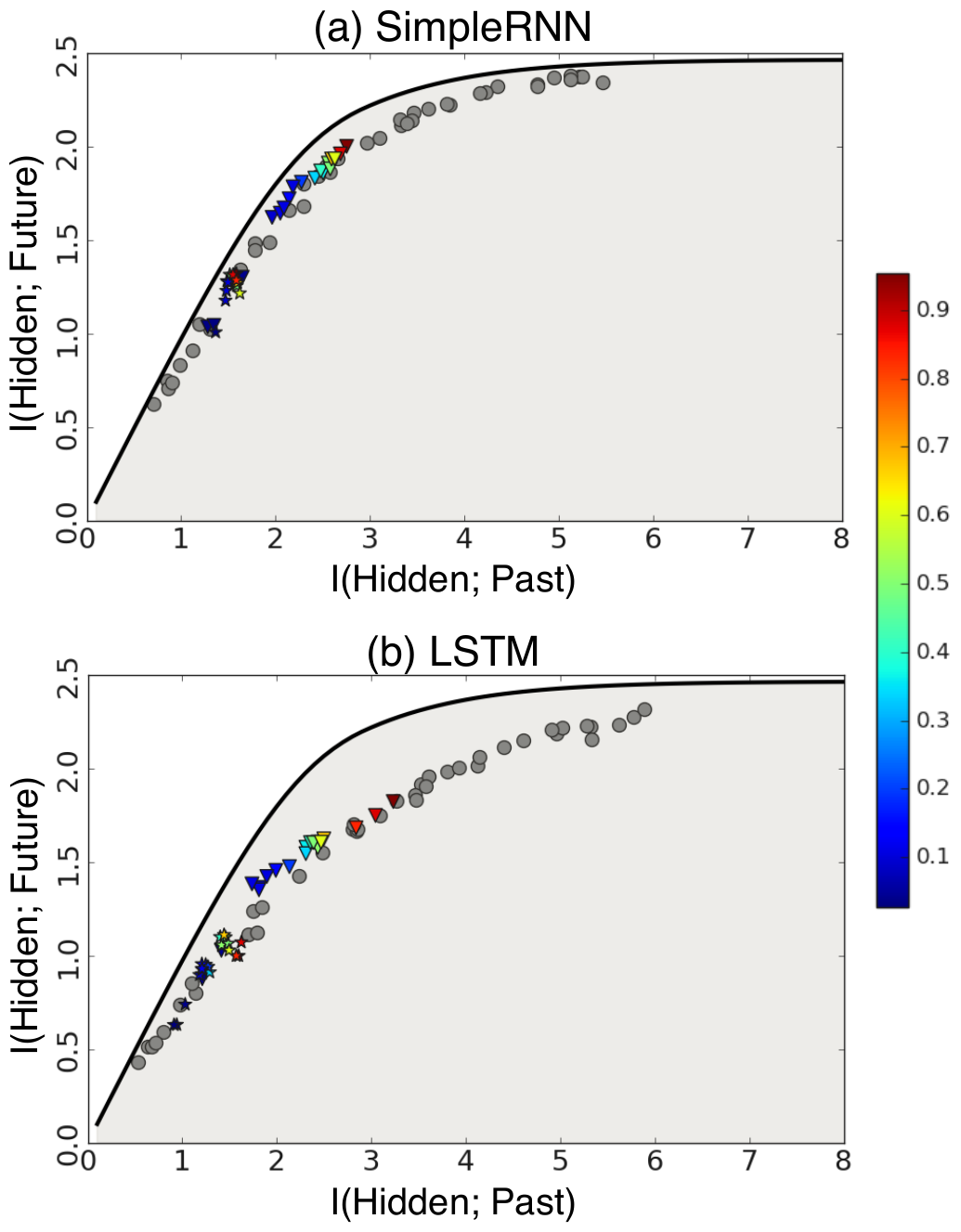}
\caption{The impact of dropout on predictive information capacity for fully connected RNNs {\bf (Top)} and LSTMs {\bf (Bottom)}. Grey $\circ$ marks the result of stochastically trained RNNs as described in~\Cref{fig:bho}. Colored marks the result for stochastically trained RNN with gaussian noise and different dropout keep probabilities on the RNN outputs, with the color determined by the keep probability (rate). $\triangledown$ markers are with noise level $0.1$, and $\star$s markers are with noise level $0.5$.}
\label{fig:app_dropout}
\end{figure}
\begin{figure}[!h]
\centering
\includegraphics[width=0.65\linewidth]{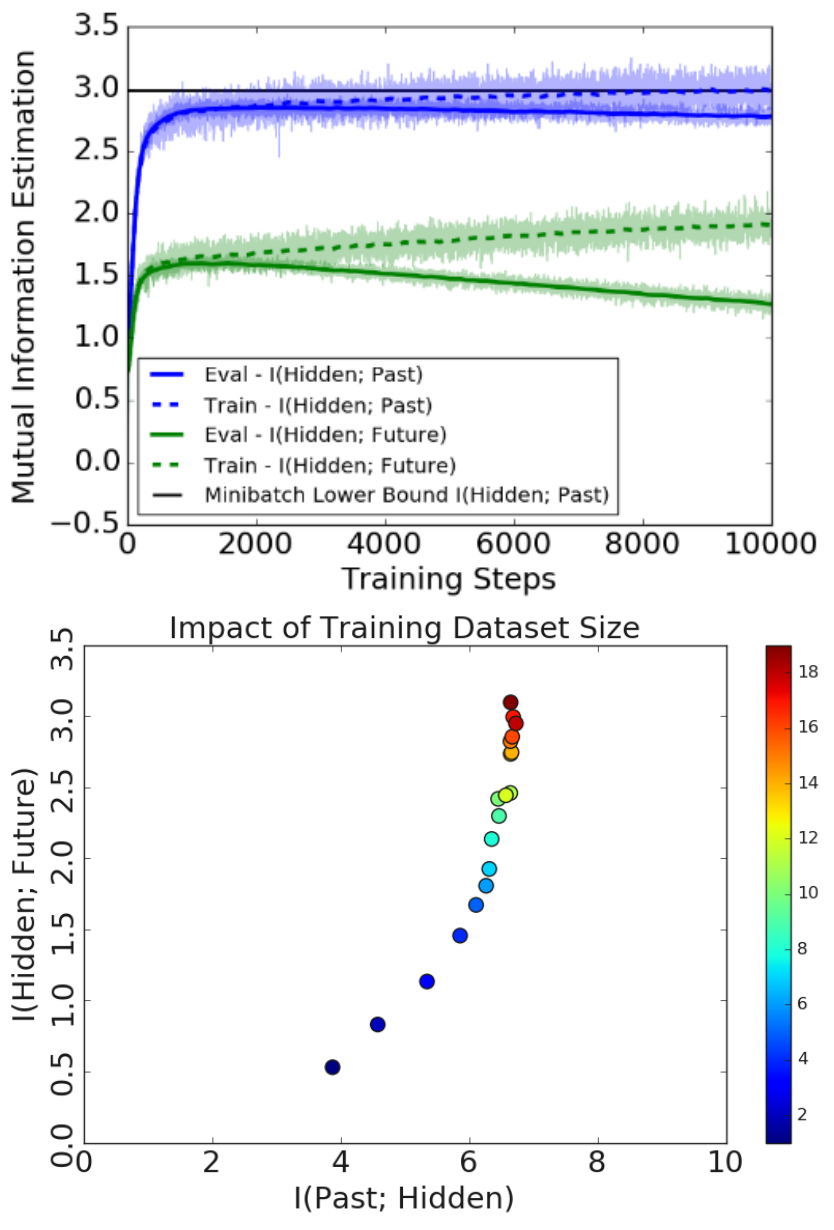}
\caption{
    {\bf (Top)} Estimates of past and future information over training iterations on the training and testing BHO data. We can see that our MI estimates quickly overfit, which we remedy here by early stopping.
    {\bf (Bottom)} Impact of training dataset sizes on InfoNCE estimator on Aaron's Sheep dataset, as introduced in~\Cref{chp_quickdraw}. The original dataset has
    7200 examples for training, and 800 for evaluation. We augment the dataset by random scaling the input values per sequence. The colors indicate the multiples of original dataset size after augmentation.
    } 
\label{fig:estimator_training}
\end{figure}

\begin{figure*}[!htb]
    \begin{adjustwidth}{-5cm}{-5cm}
      \centering
      \includegraphics[width=0.55\linewidth]{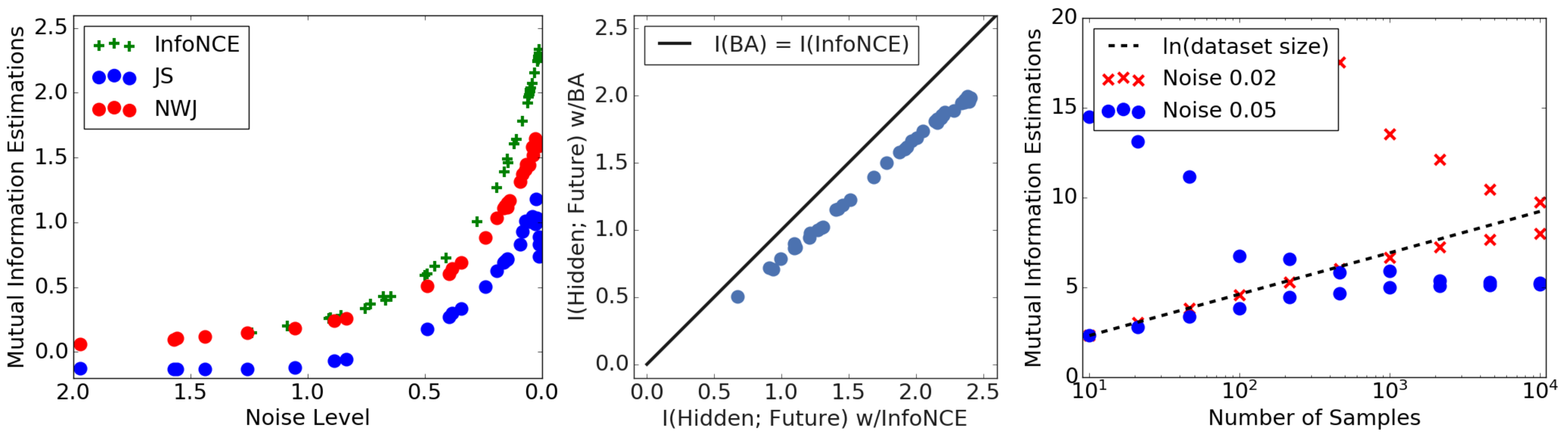}
    \end{adjustwidth}
    \caption{
    {\bf (Left)} Comparison among critic based estimators, InfoNCE, JS and NWJ. {\bf (Middle)} Comparison between estimations from Barber-Agakov and InfoNCE lower bound on future information $I(X_{\text{future}}; Z)$. {\bf (Right)} Illustration of the convergence for minibatch upper and lower bounds with two noise levels, $0.02$ (Blue) and $0.05$ (Red). Dashed line is the $\ln (\text{number of samples})$, which is the limit for minibatch lower bounds.} 
    \label{fig:bounds_comparison}
\end{figure*}

{\bf Comparison among estimators.} 
There are many different mutual information estimators. 
In \Cref{fig:bounds_comparison}, we compare various mutual information lower bounds with learned critics: InfoNCE, NWJ and
JS, as summarized in~\citet{poole2019variational}. NWJ and JS show higher variance and worse bias than InfoNCE.
The second panel of \Cref{fig:bounds_comparison} demonstrates that InfoNCE outperforms
a variational Barber-Agakov style variational lower bound at measuring the future information.
Therefore, we adopted InfoNCE as the critic based estimator for the future information in the previous
section.

For the past information, 
we could generate both tractable
upper and lower bounds,
given our tractable likelihood, $p(z_t | x_{\le t})\sim \gN(h_t, \sigma^2)$.
In the third panel of \Cref{fig:bounds_comparison} we
demonstrate that these bounds become tight as
the sample size increases.
However they require a large number of samples
before they converge.  Fundamentally, the lower bound itself is upper-bounded by the 
log of the number of samples used, requiring sample sizes exponential in the true MI to form accurate estimates.

{\bf Estimator training with finite dataset.} 
Training the learned critic on finite datasets for a large number of iterations resulted in problematic memorization and overestimates of MI. To counteract the overfitting, we performed early stopping using the MI estimate with the learned critic on a validation set. Unlike the training MI, this is a valid lower bound on the true MI. We then report estimates of mutual information using the learned critic on an independent test set, as in~\Cref{fig:estimator_training}.

{\bf Accurate MI estimates for optimal representations.}
As a final and telling justification of the efficiency of our estimators, 
\Cref{fig:gib_theory} demonstrates that our estimators
 match the true mutual information analytically derived for the optimal projections. Background for the Gaussian Information Bottleneck is
in \Cref{sec:gib} and details of the calculation can be found in \Cref{app:gib}.

{\bf Impact of training objective.}
\label{app:mle_cpc}

To assess whether the observed inefficiency in
the information plane was due to 
the maximum likelihood (MLE) objective itself,
we additionally trained contrastive predictive
coding (CPC) models~\citep{cpc}.
We used the identical model architecture described in~\Cref{sec:stochastic_rnn}, and
used the InfoNCE lower bound on the mutual information between the current time step and $K$ steps into the future to train the RNN. 
For our experiments on the Brownian harmonic oscillator, 
we look from $K=1$ to $K=30$ steps into the future, and use a linear readout from the hidden states of the RNN to a time-independent embedding of the inputs.
As shown in~\Cref{fig:app_cpc_vs_mle}, we found that models trained with InfoNCE had similar frontiers to those trained with MLE.
Thus for this dataset and architecture, the loss function does not appear to have a substantial impact. 
However, this may be due to the BHO dataset having Markovian dynamics, thus optimizing for one-step-ahead prediction with MLE is sufficient to maximize mutual information with the future of the sequence. For non-Markovian sequences, we expect that InfoNCE-trained models may be more efficient than MLE-trained models.

{\bf Comparison of dropout vs. Gaussian noise.}
\label{app:dropout}
Dropout~\citep{srivastava2014dropout} is a common method applied on neural network training to prevent overfitting and a potential alternative way to eliminate information. We trained fully connected RNNs and LSTMs with different levels of dropout rate. 
As shown in~\Cref{fig:app_dropout}, 
we find that RNNs trained with dropout extract less information than the ones without it, but the information frontier of the models does not change,
when we sweep dropout rate and additive noise. 
We also demonstrate that 
our simple noise injection technique
can find equivalent models.

\subsection{Are RNNs efficient on real-world datasets?}
\label{chp_quickdraw}

While we have found that deterministically trained RNNs are inefficient in the information plane on the BHO dataset, it is not clear whether this is an intrinsic property of RNNs or specific to that particular synthetic dataset. To assess whether the same inefficiencies were present on real-world datasets, we performed additional experiments on two hand-drawn sketch datasets that are similar in structure and dimensionality but have more complicated non-Markovian structure. 
The sketch datasets we consider consist of a sequence of tuples $(x, y, p)$ denoting the $(x, y)$ position of the pen,
as well as a binary $p$ 
denoting whether the pen is up 
or down.
The first set of experiments analyzed the Aaron Koblin Sheep Sketch Dataset\footnote{Available from \url{https://github.com/hardmaru/sketch-rnn-datasets/tree/master/aaron\_sheep}}. Full experimental
details are in
\Cref{app:quickdraw_training_details}.
The RNN architecture we used is based on the decoder RNN component of SketchRNN, trained with MLE (optionally injecting noise) and  online data augmentation as in
\citet{ha2017sketchrnn}.

\begin{figure}[!htb]
\centering
\includegraphics[width=\linewidth]{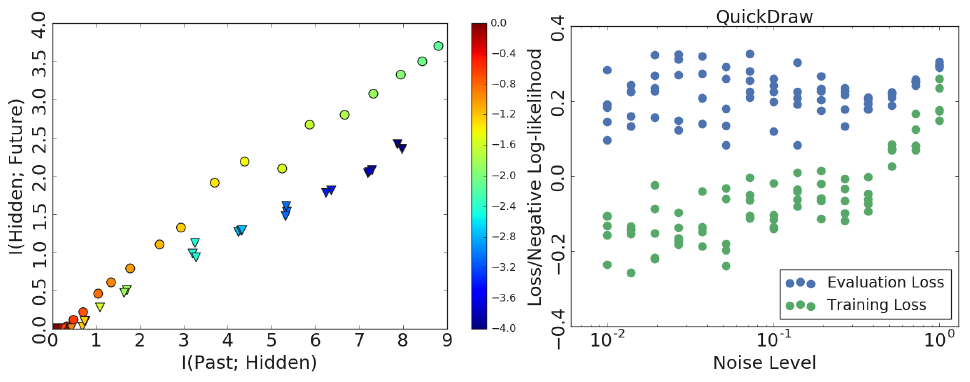}
\caption{{\bf (Left)} Evaluation on Aaron Sheep dataset by comparing training explicitly with noise ($\circ$) and post-hoc noise injection after training ($\triangledown$). The color bar shows the noise level in $\log_{10}$ scale. 
{\bf (Right)} Comparing the training and evaluation loss for noise-trained RNNs.}
\label{fig:quickdraw}
\end{figure}

{\bf Information-constrained RNNs are more efficient.}
First, we performed the same set of experiments as in \Cref{sec:bho_training}, training deterministic
and information-constrained RNNs by adding noise to the hidden state. We then estimated past and future information using InfoNCE and minibatch lower bound, respectively. \Cref{fig:quickdraw} (left) shows the estimates on the information plane for the trained networks, similar to \Cref{fig:bho}. Again, the networks that were trained with information constraints (circular markers) instead of evaluated post-hoc with noise (triangular markers) dominate in the information plane. 
For this natural dataset, we no longer know the optimal frontier on the information plane, but still see that the deterministically trained networks evaluated with noise are suboptimal compared to the simple stochastic networks trained with noise.

RNNs trained with higher levels of compression -- achieved through higher levels of injected noise -- obtained similar performance to deterministically-trained networks in terms of heldout likelihood but with lower variance across runs
(\Cref{fig:quickdraw}(right)). 
This provides preliminary evidence that for large datasets, information constraints can still be useful for reproducibility, not just compression.

{\bf Compression improves heldout likelihood and conditional generation in the data-limited regime.}

\begin{figure}[!htb]
\centering
\includegraphics[width=\linewidth]{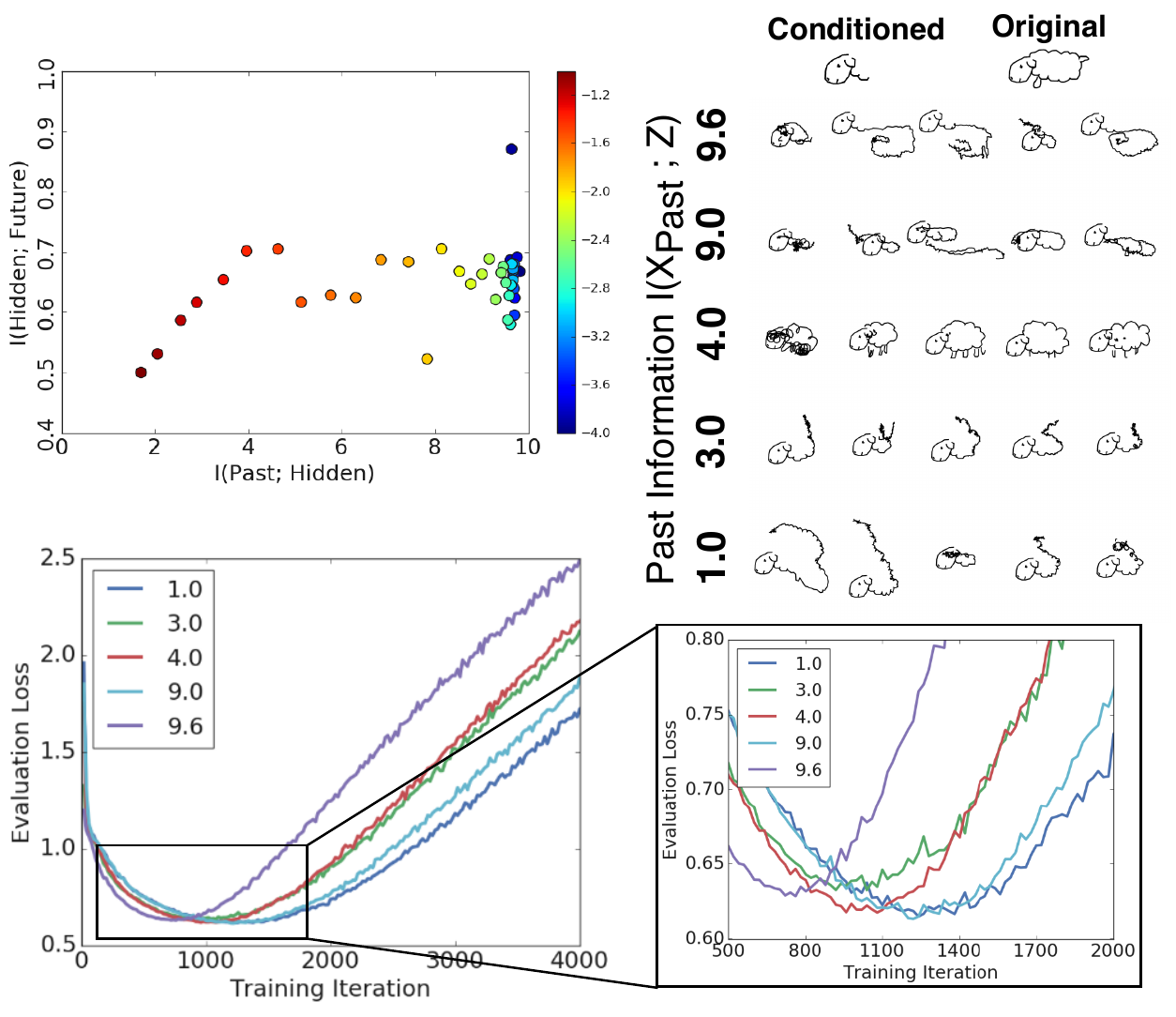}
\caption{{\bf Top left:} Estimation of past and future information for RNN trained with $100$ samples from the sheep sketch dataset, with color indicating the level of noise on a $\log_{10}$ scale. {\bf Bottom:} Validation loss for different noise levels. {\bf Top right:} Conditional generated samples from models with different levels of past information. The generation is conditioned on a $25$-step stroke, which is taken from a held-out sample. The samples from the model with $4.0$ nats past information is qualitatively of better sample quality than models with higher past information. For additional samples, see \Cref{fig:appendix_samples}.}
\label{fig:quickdraw_limited_sample}
\end{figure}

We expect the benefits of compressed representations to be most noticeable in the data-limited regime where compression may act as a regularizer to prevent the network from overfitting to a small training set. 
As the training set size increases, such regularization may be less effective for learning a good model even when it aids in compression of the hidden state.

To investigate the impact of constraining past information in RNNs in the data-limited regime, we repeated the experiments of the previous section with a limited dataset of only $100$ examples (vs. the 1000 examples used previously).
\Cref{fig:quickdraw_limited_sample} (top left) shows the corresponding information plane points for stochastically trained RNNs with various noise levels.  Notice that at about 4 nats of past information,
the networks future information essentially saturates. 
While it seems as though the networks 
do not suffer reduced performance even when learning higher capacity representations, 
this appears to be due to early stopping which was included in the training procedure. 
As can be seen in~\Cref{fig:quickdraw_limited_sample} (lower left), 
all of our networks overfit in terms of evaluation loss, but the onset of overfitting was strongly controlled by the degree of compression.
Most noticeably, in the limited data regime, 
compressed representations lead to improved sample quality, 
as seen in~\Cref{fig:quickdraw_limited_sample} (right).
Models with intermediately-sized compressed representations show the
best generated samples while retaining a good amount of diversity.  
Models with either too little
or too much past information tend to produce nonsensical generations.

\subsection{Is compression useful for downstream tasks on the QuickDraw Sketch Dataset?}
\begin{figure}[htb]
\centering
\includegraphics[width=0.9\linewidth]{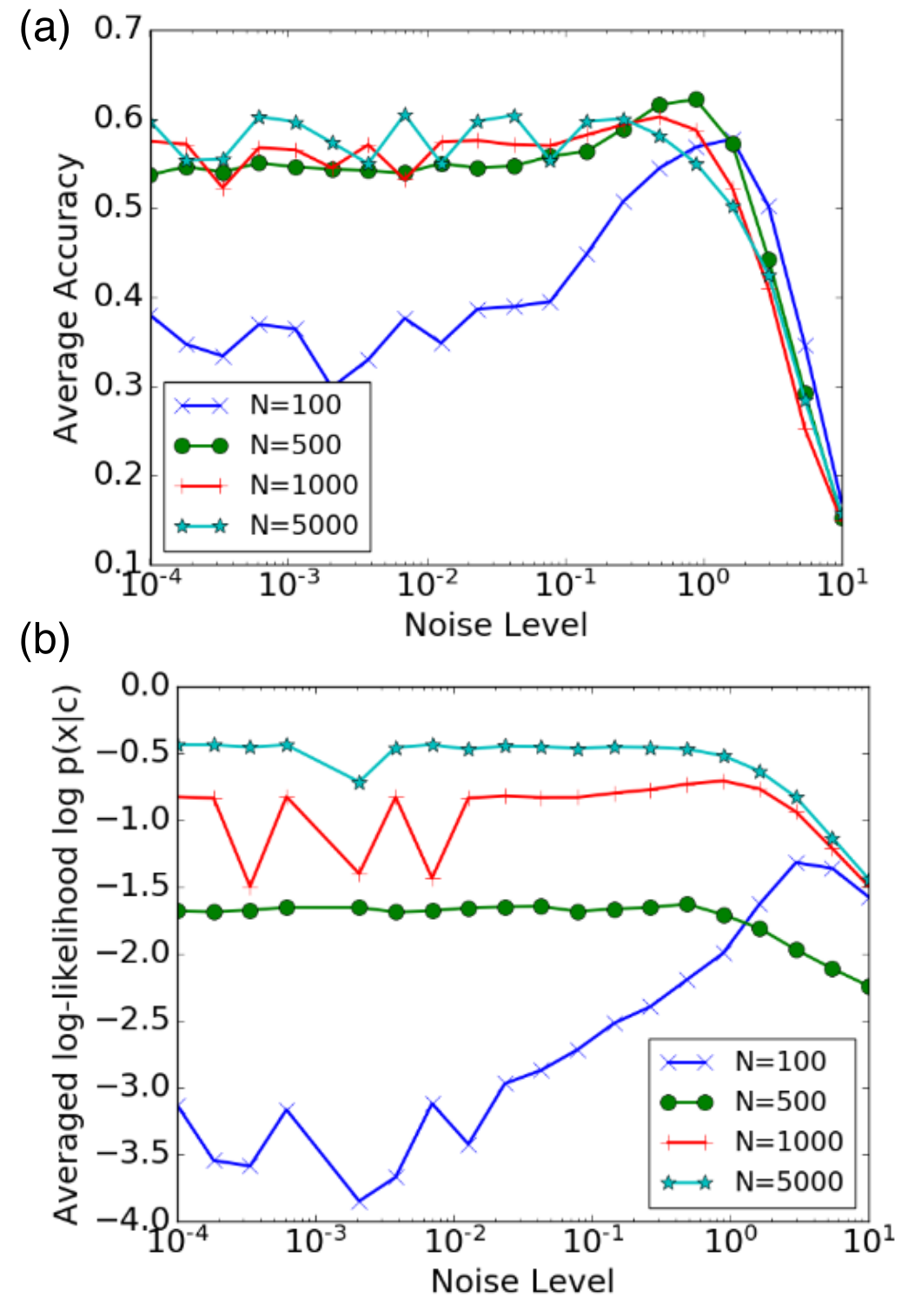}
\caption{
{\bf (a)} The averaged accuracy of the classifiers. 
{\bf (b)} The averaged posterior log-likelihoods of selecting the correct class labels using the naive Bayes classifiers trained with different noise levels, $\log p(c|x) = \log \left(p (x | c )  p(c)/ p(x) \right)$.
}
\label{fig:quickdraw_classification}
\end{figure}

To further evaluate the utility of constraining information in RNNs, we experimented on a real-world classification task using the QuickDraw sketch dataset \citep{quickdraw} (which is distinct from the sheep dataset used in the previous section). This dataset consists of hand-drawn sketches where a subject was asked to draw a particular class in a time-constrained setting, resulting in diverse sketches represented as sequences of pen strokes. We formed a dataset containing examples from 11 classes: apple, donut, flower, hand, leaf, pants, sheep, van, camel, shorts, and pear, and constructed training sets of various sizes (from 100 to 5,000 examples per class) and a test set with 2500 examples per class.
These distinctly-sized training sets were used to assess the interaction of dataset size with information constraints. For each class, dataset-size and level of noise (corresponding to different information constraints), we trained a class-conditional RNN, producing a class-conditional generative model $p(x|c)$. 
We evaluated these class conditional RNNs in two ways: (1) average heldout log-likelihood (as in \Cref{chp_quickdraw}), and (2) accuracy when used in a Na\"ive Bayes classifier.

{\bf Constraining information improves heldout likelihood for small datasets.}
In \Cref{fig:quickdraw_classification}(b), we plot the average test log-likelihood as a function of noise level, averaging across the 11 classes. We find that for small datasets sizes (100 examples per class), adding more noise to the hidden state, and thus lowering the amount of information extracted about the past, improves the test log-likelihood. As the dataset size increases, we see less benefit in constraining information. 
However, we can see that even up to 5000 examples per class, we can greatly reduce the amount of information stored in the hidden state without any noticeable drop in heldout likelihood. In other words, we can greatly compress the RNN hidden state with no loss in performance.

{\bf Constraining information improves classification accuracy.}
To evaluate the impact of limiting past information on downstream classification tasks, we constructed a simple Na\"ive Bayes classifier from the class-conditional RNNs. Given an input $x$, we can compute the posterior distribution over classes as: $p(C=j | x) = \frac{p(x | C=j) p(C=j)}{\sum_k p(x | C=k) p(C=k)}$. Here we assume a uniform probability over classes, and thus can compute the predicted class distribution by evaluating each class-conditional RNN. 
We can then evaluate accuracy by checking whether the $\argmax$ over $p(c|x)$ is equal to the true class for each point in the test set. In \Cref{fig:quickdraw_classification}(a), we plot the average classification accuracy as a function of noise level, finding that for most dataset sizes classification accuracy improves by constraining information. As the dataset size increases, we can see that the best accuracies are achieved by smaller amounts of noise, indicating that regularization through information constraints may only be beneficial for downstream classification when the dataset size is limited.

\section{Discussion}
In this work, we have demonstrated how analyzing RNNs in terms of predictive information can be a useful tool for probing and understanding behavior. We find that deterministically-trained RNNs are inefficient, extracting more information about the past than is required to predict the future. By analyzing different training objectives and noise injection approaches in the information plane, we can better understand the tradeoffs made by different models, and identify models that are closer to the optimality demonstrated by biological neurons \citep{palmer2015predictive}.

While the simple strategy of adding noise to a bounded hidden state can be used to constrain information, setting the amount of noise and identifying where one should be on the information plane remains an open problem. Additionally, studying the impact of learning objectives, optimization choices like early stopping, and other architecture choices, such as stochastic latent variables in variational RNNs \citep{chung2015recurrent}, or attention-based Transformers \citep{vaswani2017attention} in the information plane could yield insights into their improved performance on several tasks.

Finally, the impact of constraining information on model performance and downstream tasks largely remains an open problem. When should we constrain information and for which tasks is compression useful? Our preliminary results indicate that constraining information can improve downstream classification performance for simple sketch datasets, but many models have demonstrated excellent performance through information maximization alone without information constraints \citep{cpc, hjelm2018learning}.

\bibliography{reference}
\bibliographystyle{icml2020}

\clearpage
\appendix
\section{Appendix}
\subsection{Gaussian Information Bottlenecks}
\label{app:gib}

Consider jointly multivariate Gaussian random variables $X \in \R^{D_X}$ and $Y \in \R^{D_Y}$, with covariance $\Sigma_X$ and $\Sigma_Y$ and cross-covariance $\Sigma_{XY}$.
The solution to the Information Bottleneck objective:
\begin{equation}
    \min_{T} I(X;T) - \beta I(Y;T),
\end{equation}
is given by a linear transformation $T = \mA X + \varepsilon$ with $\varepsilon \sim \gN(\mathbf{0}, \Sigma_\varepsilon)$.
The projection matrix $A$ projects along
the lowest eigenvectors $\lambda_i (i\in [1, D_X])$ of $\Sigma_{X|Y} \Sigma_X^{-1}$, where the trade-off parameter $\beta$
decides how many of the eigenvectors participate, ${\bf \vv}_i^T (i\in [1, D_X])$. 
The projection matrix $\mA$ could be analytically derived as 
\begin{equation}
    \mA = [{\bf \alpha_1 \vv_1^T, \alpha_2 \vv_2^T, \dots, \alpha_{D_X} \vv_{D_X}^T}],
    \ \Sigma_\varepsilon = \mathbb{I}
\end{equation}
where the projection coefficients ${\bf \alpha_i^2}= \text{max}(\frac{\beta(1-\lambda_i)-1}{\lambda_i r_i}, 0),~r_i=\bf{\vv}_i^T\Sigma_X \bf{\vv}_i$, with proof in \Cref{appendix:optimal_projection}.

Given the optimally projected states $T$, the \emph{optimal frontier} (black curve in \Cref{fig:gib_theory}) is:
\begin{eqnarray}
& I(T;Y) = I(T;X)\\
&- \frac{n_I}{2} \log(\prod\limits_{i=1}^{n_I}(1-\lambda_i)^{\frac{1}{n_I}} + e^{\frac{2I(T;X)}{n_I}}\prod\limits_{i=1}^{n_I}\lambda_i^{\frac{1}{n_I}}),\\
& c_{n_I} \le I(T;X)\le c_{n_I+1}
\label{eqn:gib_segments}
\end{eqnarray}
where $n_I$ is the cutoff number indicating the number of smallest eigenvalues being used. The critical points $c_{n_I}$, changing from using $n_I = N$ eigenvalues to $N+1$ eigenvalues, can be derived given the concave and $C^1$ smoothness property for the optimal frontier, with proof in \Cref{appendix:information_frontier}:
\begin{equation}
c_{n_I} = \frac{1}{2}\sum\limits_{i=1}^N \log\frac{\lambda_{N+1}}{\lambda_i}
\frac{1-\lambda_i}{1-\lambda_{N+1}}
\end{equation}

\subsubsection{Proof of Optimal Projection}
\label{appendix:optimal_projection}
By Theorem~3.1 of \cite{chechik2005gib}, the projection matrix for optimal projection is given by 
\begin{equation}
    \mA = 
        \left\{
                \begin{array}{c l}
                    \left [{\bf \alpha_1 \vv_1^T, 0,  \dots, 0} \right], & 0 \le \beta \le \beta_1 \\
                    \left [ {\bf \alpha_1 \vv_1^T, \alpha_2 \vv_2^T,  \dots, 0} \right], & \beta_1 \le \beta \le \beta_2  \\
                    \vdots & 
                \end{array}
         \right\}
\label{eqn:app_proj_a}
\end{equation}
where ${\bf \vv}_i^T (i\in [1, D_X])$ are left eigenvectors of $\Sigma_{X|Y} \Sigma_X^{-1}$ sorted in ascending order by the eigenvalues $\lambda_i (i\in [1, D_X])$; $\beta_i = \frac{1}{1-\lambda_i}$ are critical values for trade-off parameter $\beta$; and the projection coefficients are ${\bf \alpha_i^2}= \frac{\beta(1-\lambda_i)-1}{\lambda_i r_i}$, $r_i=\bf{\vv}_i^T\Sigma_X \bf{\vv}_i$. 
In practice, noticing that $\beta * (1-\lambda_i) - 1 < 0$ when $\beta < \beta_i$, we simplify \Cref{eqn:app_proj_a} as 
$\mA = [{\bf \alpha_1 \vv_1^T, \alpha_2 \vv_2^T, \dots, \alpha_{D_X} \vv_{D_X}^T}]$
with ${\bf \alpha_i^2}= \text{max}(\frac{\beta(1-\lambda_i)-1}{\lambda_i r_i}, 0)$.

\subsubsection{Proof of Critical Points on Optimal Frontier}
\label{appendix:information_frontier}
By Eq.15 of \cite{chechik2005gib}
\begin{align*}
& I(T;Y) = I(T;X)\\
&- \frac{n_I}{2} \log(\prod\limits_{i=1}^{n_I}(1-\lambda_i)^{\frac{1}{n_I}} + e^{\frac{2I(T;X)}{n_I}}\prod\limits_{i=1}^{n_I}\lambda_i^{\frac{1}{n_I}})
\end{align*}
where $n_I$ is the cutoff on the number of eigenvalues used to compute the bound segment, with eigenvalues sorted in ascending order.

In order to calculate the changing point, where one switching from choosing $n_I = N$ to $N+1$, 
by $C^1$ smoothness conditions:
\begin{equation}
\frac{\mathrm{d} I_{n_I=N}(T;Y)}{\mathrm{d} I(T;X)} = \frac{\mathrm{d} I_{n_I=N+1}(T;Y)}{\mathrm{d} I(T;X)}
\label{eqn:app_smoothness}
\end{equation}

LHS is
\begin{eqnarray}
L.H.S. & =&  1-\frac{\mathrm{d} I_{n_I=N}(T;Y)}{\mathrm{d} I(T;X)} \\
&=& 
\frac{\prod\limits_{i=1}^{N}(\lambda_i)^{\frac{1}{N}}e^{\frac{2I(T;X)}{N}}}
{\prod\limits_{i=1}^{N}(1-\lambda_i)^{\frac{1}{N}} + e^{\frac{2I(T;X)}{n_I}}\prod\limits_{i=1}^{N}\lambda_i^{\frac{1}{N}}} \\
&=& 
\frac{e^{\frac{2I(T;X)}{N}}}{e^{\frac{2I(T;X)}{N}}+\prod\limits_{i=1}^{N}(\frac{1-\lambda_i}{\lambda_i})^{\frac{1}{N}}} 
\end{eqnarray}

Thus, \Cref{eqn:app_smoothness} could be rewritten as
\begin{align*}
& \frac{e^{\frac{2I(T;X)}{N}}}{e^{\frac{2I(T;X)}{N}}+\prod\limits_{i=1}^{N}(\frac{1-\lambda_i}{\lambda_i})^{\frac{1}{N}}} 
\\
& =
\frac{e^{\frac{2I(T;X)}{N+1}}}{e^{\frac{2I(T;X)}{N+1}}+\prod\limits_{i=1}^{N+1}(\frac{1-\lambda_i}{\lambda_i})^{\frac{1}{N+1}}} 
\end{align*}

Rewrite RHS of above equation, and noticing $\frac{1}{n(n+1)}=\frac{1}{n} - \frac{1}{n+1}$
\begin{align*}
R.H.S. &= \frac{e^{\frac{2I(T;X)}{N+1}}}{e^{\frac{2I(T;X)}{N+1}}+\prod\limits_{i=1}^{N+1}(\frac{1-\lambda_i}{\lambda_i})^{\frac{1}{N+1}}} \\
&
= \frac{e^{\frac{2I(T;X)}{N}}}{e^{\frac{2I(T;X)}{N}}+e^{\frac{2I(T;X)}{N(N+1)}}\prod\limits_{i=1}^{N+1}(\frac{1-\lambda_i}{\lambda_i})^{\frac{1}{N+1}}} 
\end{align*}

The term in lower right corner could be written as
\begin{align*}
&\prod\limits_{i=1}^{N+1}(\frac{1-\lambda_i}{\lambda_i})^{\frac{1}{N+1}} \\
&= \prod\limits_{i=1}^N(\frac{1-\lambda_i}{\lambda_i})^{\frac{1}{N+1}}
(\frac{1-\lambda_{N+1}}{\lambda_{N+1}})^{\frac{1}{N+1}} \\
&= \left(\prod\limits_{i=1}^N(\frac{1-\lambda_i}{\lambda_i})^{\frac{1}{N}} \right)
\left(\prod\limits_{i=1}^N\left(\frac{\lambda_i(1-\lambda_{N+1})}{(1-\lambda_i)\lambda_{N+1}}\right)^{\frac{1}{N(N+1)}}\right)
\end{align*}


To let LHS = RHS, one trivial solution is
\begin{equation}
\mathrm{1}
=
e^{\frac{2I(T;X)}{N(N+1)}}
\left(\prod\limits_{i=1}^N\left(\frac{\lambda_i(1-\lambda_{N+1})}{(1-\lambda_i)\lambda_{N+1}}\right)^{\frac{1}{N(N+1)}}\right)
\end{equation}

Taking $\log$ and cancelling out multiplicative factors, one get the critic point to change from $n_i = N$ to $n_i = N+1$ happens at 
\begin{equation}
I(T;X) = \frac{1}{2}\sum\limits_{i=1}^N \log\frac{\lambda_{N+1}}{\lambda_i}
\frac{1-\lambda_i}{1-\lambda_{N+1}}
\end{equation}

The original result written in \cite{chechik2005gib} is missing a factor of $\frac{1}{2}$.

\subsubsection{Optimal Projection}
\begin{figure}[htb]
    \begin{adjustwidth}{-5cm}{-5cm}
    \centering
    \includegraphics[width=0.4\linewidth]{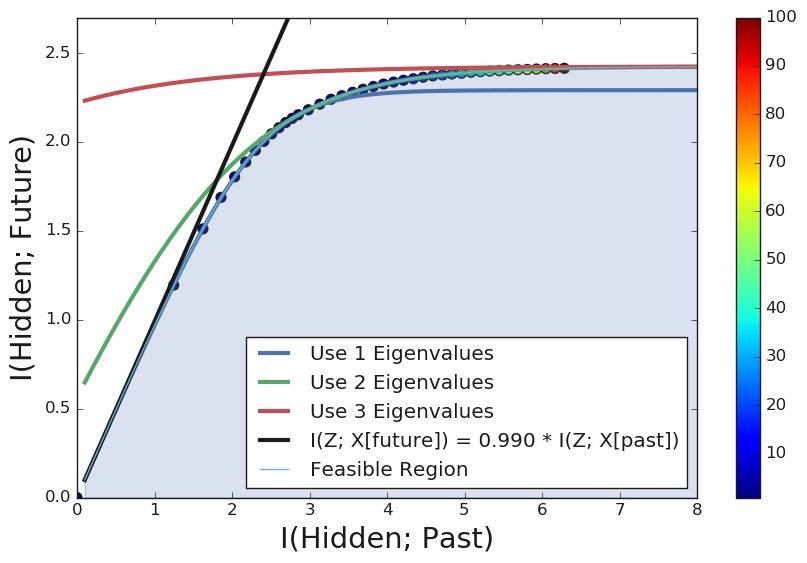}
    \end{adjustwidth}
    \caption{Conditionally generated samples from models with different levels of past information.} 
\label{fig:appendix_optimal_frontier}
\end{figure}

The optimal frontier is generated by joining segments described by \Cref{eqn:gib_segments}, as illustrated in \Cref{fig:appendix_optimal_frontier}. 

\subsection{Details for Brownian Harmonic Oscillator}
\label{app:bho_training_details}
To generate the sample trajectories,
we set the undamped angular velocity $\omega = 1.5 \times 2 \pi (\text{rad})$,  
damping coefficient $\Gamma = 20.0$, 
and the dynamical range of external forces $D=1000.0$, with integration time step-size $\Delta t = 0.01667$.
The stationary distribution of \Cref{eqn:bho_eom} is analytically derived in~\citet{PRE_Brownian}.

We train RNNs with infinite number of training samples, which are generated online and divided into batches of $32$ sequences. RNNs, including fully connected RNN, GRU and LSTM, are all with $32$ hidden units and {\tt tanh} activation. They are trained with momentum optimizer~\citep{momentum} for $20000$ steps, with $\text{momentum} = 0.9$ and gradient norm being clipped at $5.0$. Learning rate for training is exponentially decayed in a stair-case fashion, with initial learning rate $10^{-4}$, decay rate $0.9$ and decay steps $2000$. 

The mutual information estimators, with learned critics, are trained for $200000$ steps with Adam optimizer~\citep{Kingma2015adam} at a flat learning rate of $10^{-3}$. The training batch size is $256$, and the validation and evaluation batch sizes are $2048$. We use early stopping to deal with overfitting. The training is stopped when the estimation on validation set does not improve for $10000$ steps, or when it drops by $3.0$ from its highest level, whichever comes first. We use separable critics~\citep{poole2019variational} for training the estimators. Each of the critics is a three-layer MLP, with $[256, 256, 32]$ hidden units and [ReLU, ReLU, None] activations. The weights for each layer are initialized with Glorot uniform initializer~\citep{glorot2010initliazer}, and the biases are with He normal initializer~\citep{he2015initializer}. For the minibatch upper and lower bounds, they are estimated on batches of $4096$ sequences.

To train the critics, we feed $100$-step BHO sequences into trained RNN to get RNN hidden states and conditional distribution parameters. From each sequence, we use last $36$ steps for the inputs to the estimators, where first $18$ steps as $x_\mathrm{past}[t],\ t=[1, 2, \dots 18]$, and the other $18$ steps as $x_\mathrm{future}[t],\ t=[19, 20, \dots 36]$. The hidde state $z_{18}$ is extracted at the last time step of $x_\mathrm{past}$.

\subsection{Training Details for Vector Drawing Dataset}
\label{app:quickdraw_training_details}

We train decoder-only SketchRNN~\citep{ha2017sketchrnn} on Aaron Koblin Sheep Dataset, as provided in \url{https://github.com/hardmaru/sketch-rnn-datasets/tree/master/aaron_sheep}. The SketchRNN uses LSTM as its RNN cell, with $512$ hidden units. 

For RNN training, We adopt the identical hyper-parameters as in \url{https://github.com/tensorflow/magenta/blob/master/magenta/models/sketch_rnn/model.py}, except that we turn off the recurrent drop-out, since drop-out masks out informations and will interfere with noise injection. 

For mutual information estimations, we use the identical hyper-parameters as descibed in \Cref{app:bho_training_details}, except that: the evaluation batch size for critic based estimator, InfoNCE, is set to be $4096$, and $16384$ for minibatch bounds; early stopping criteria are changed to that either the estimation does not improve for $20000$ steps or drops by $10.0$ from its highest level, whichever comes first.

Due to the limitation of the sequence length of Aaron's Sheep, we use the samples with at least $36$ steps long. The $x_\mathrm{past}$ and $x_\mathrm{future}$ are split at the middle of the sequences, and each with $18$ steps. 

Due to the limitation of the dataset size of Aaron's Sheep, we augment the dataset with randomly scale the stroke by a factor sampled from $\gN(0, 0.15)$ for each sequence to generate a large dataset. \Cref{fig:estimator_training} (Right) shows that the augmentation helps in training the estimator.

\begin{figure*}[htb]
    \centering
      \includegraphics[width=0.85\linewidth]{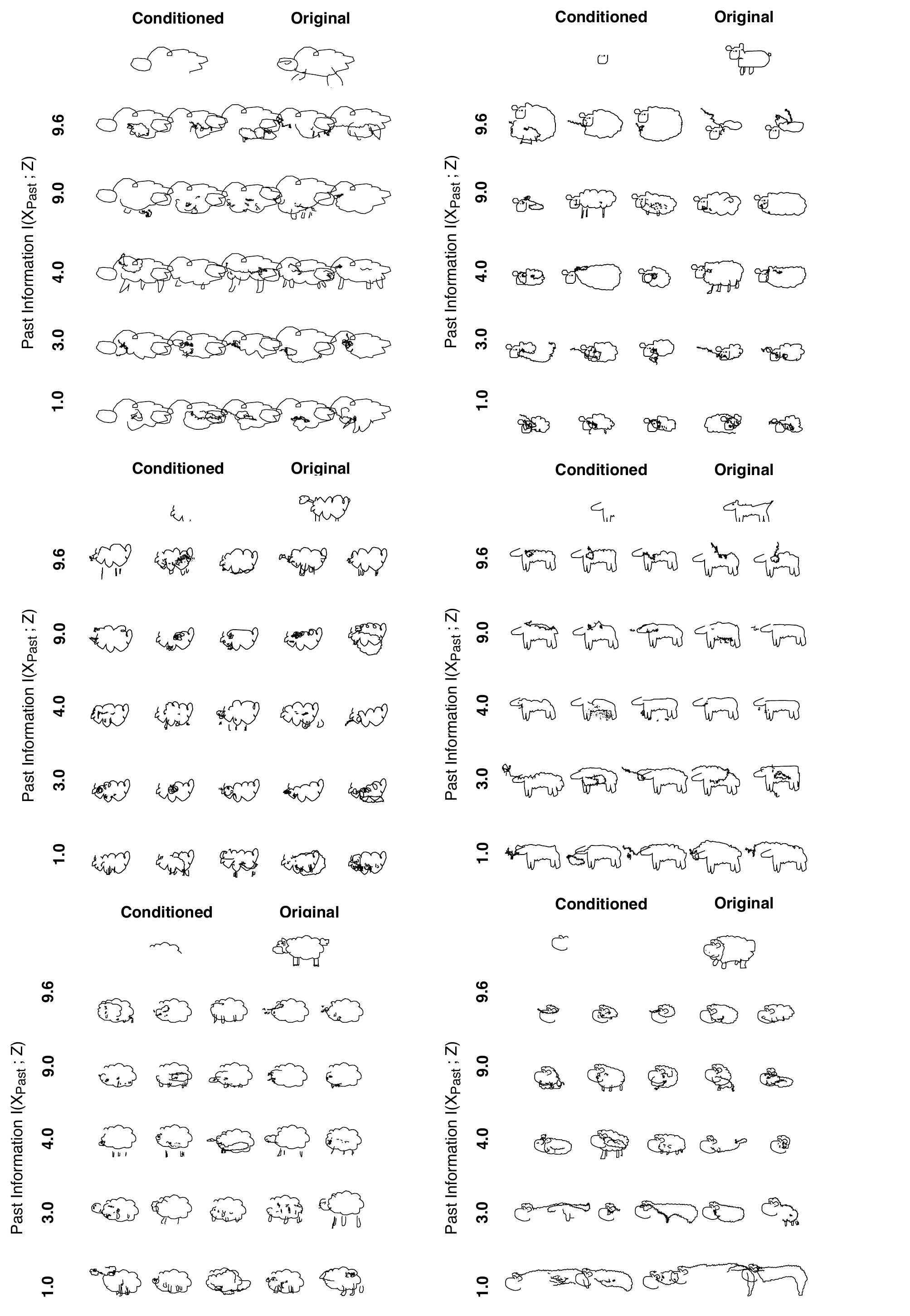}
    \caption{Conditionally generated samples from models with different levels of past information.} 
\label{fig:appendix_samples}
\end{figure*}

\end{document}